%% file: main.tex
\documentclass[11pt]{article}

\usepackage[T1]{fontenc}
\usepackage[utf8]{inputenc}
\usepackage{lmodern}
\usepackage[margin=1in]{geometry}
\usepackage{microtype}
\usepackage{amsmath,amssymb}
\usepackage{booktabs}
\usepackage{tabularx}
\usepackage{array}
\usepackage{enumitem}
\usepackage{xcolor}
\usepackage{graphicx}
\usepackage{tikz}
\usetikzlibrary{arrows.meta,positioning,fit,calc}
\usepackage[round,authoryear]{natbib}
\usepackage{xurl}
\usepackage[hidelinks]{hyperref}
\hypersetup{
  pdftitle={Artificial Intelligence in Equity and Crypto Markets: Progress, Profitability Evidence, and the Limits of Automated Investing},
  pdfauthor={Linsen Zhu and Mengqing Cai},
  pdfsubject={A critical state-of-the-art review of AI in equity and crypto markets},
  pdfkeywords={artificial intelligence, machine learning, asset pricing, algorithmic trading, large language models, reinforcement learning, cryptocurrency, financial agents}
}

\newcolumntype{Y}{>{\raggedright\arraybackslash}X}

\title{Artificial Intelligence in Equity and Crypto Markets:\\
Progress, Profitability Evidence, and the Limits of Automated Investing}
\author{Linsen Zhu \and Mengqing Cai}
\date{Literature cutoff: 31 August 2026}

\begin{document}
\maketitle

\input{sections/00_abstract}
\input{sections/01_introduction}
\input{sections/02_scope_method}
\input{sections/03_alpha_translation}
\input{sections/04_prediction_representation}
\input{sections/05_text_multimodal_agents}
\input{sections/06_portfolio_rl_risk}
\input{sections/07_market_strata_execution}
\input{sections/08_profitability_evidence}
\input{sections/09_research_agenda}
\input{sections/10_conclusion}
\input{sections/11_ai_disclosure}

\bibliographystyle{plainnat}
\bibliography{references}

\end{document}

%% file: sections/00_abstract.tex
\begin{abstract}
Artificial intelligence (AI) now supports investment workflows from data and prediction through research, portfolios, execution, and tool use. Technical capability, however, is not evidence of investment profitability. This critical state-of-the-art review examines public research available through 31 August 2026 on listed equities, exchange-traded funds, centralized crypto spot, perpetual futures, and on-chain markets. We organize evidence with an \emph{alpha-translation chain}: point-in-time information must yield a stable signal, feasible positions, executable orders, and risk-adjusted returns after costs. Across machine learning, time-series foundation models, financial language models, reinforcement learning, and agents, the examined record shows real but mainly upstream progress in prediction, text processing, portfolio design, and workflow integration. Evidence is thinner for durable net performance. Temporal contamination, repeated selection, survivorship, weak benchmarks, implementation costs, venue mechanics, and capacity can break translation to net alpha. Strong historical results coexist with predictor decay, corrected look-ahead failures, mixed prospective evidence, and few audited live-capital records. Crypto adds informative state but requires separate treatment of spot, perpetual, and decentralized cash flows and execution. Within the public evidence examined here, no general AI architecture is shown to deliver persistent, cross-regime, capacity-aware net alpha. More credible claims require point-in-time data and models, decision-aligned objectives, joint portfolio--execution evaluation, controlled adaptation, prospective tests, and authority-matched governance. These conditions can improve evidence and implementation; they do not guarantee profit.
\end{abstract}


%% file: sections/01_introduction.tex
\section{Introduction}

Artificial intelligence (AI) has moved from a specialized forecasting technique to an organizing layer for investment workflows. A contemporary system may ingest prices, accounting variables, analyst text, news, social media and blockchain state; generate forecasts or research narratives; translate them into holdings under risk constraints; and route orders through broker, exchange or smart-contract tools. The literature mirrors this expansion. Cross-sectional machine learning (ML) studies estimate nonlinear return functions from hundreds of firm characteristics \citep{Gu2020EmpiricalAssetPricing,Chen2024DeepAssetPricing}. Deep policies map state variables directly to portfolio weights \citep{Simon2026DeepPolicies}; cost-aware methods optimize implementable portfolios rather than forecasts \citep{Jensen2026ImplementableFrontier}; and natural language processing (NLP) turns unstructured disclosure and news into investable signals \citep{Tetlock2007Media,Loughran2011Liability,Ke2019TextReturns}. More recently, large language models (LLMs) and agents have joined retrieval, memory, debate, analysis, order APIs and feedback loops within a single interface \citep{Yu2023FinMem,Zhang2024FinAgent,Yu2024FinCon,Xiao2024TradingAgents}. These studies document genuine technical progress. They do not, by themselves, answer the economically decisive question: does the progress produce sustainable net alpha?

That question is harder than asking whether a model predicts labels or whether a backtest earns a positive return. A forecast is not a portfolio; a portfolio return is not alpha unless the risk adjustment is stated; gross alpha is not net alpha; and a historical test labelled ``out of sample'' is not a prospectively committed trading record. Even a correctly timestamped and cost-adjusted simulation can overstate attainable performance if it ignores capacity, market impact, borrowing, execution delay, model-selection multiplicity, or changes caused by other investors adopting the same signal. Conversely, the observation that many published anomalies decay does not imply that all predictability is spurious. Large open replication exercises reproduce substantial parts of the cross-sectional record \citep{Chen2022OpenSource,Jensen2023GlobalFactors}, while institutional transaction data suggest that some established styles are cheaper and more scalable than pessimistic trading-cost models imply \citep{Frazzini2018TradingCosts}. The relevant assessment is therefore neither technological enthusiasm nor blanket market-efficiency scepticism. It is a stage-by-stage audit of what survives implementation.

This review develops that audit as an \emph{alpha-translation chain}. Information must be available at the decision time; a learned representation must produce incremental and stable information; the decision rule must convert it into positions under financing and risk constraints; orders must receive plausible fills; and realized gains must remain after fees, spreads, price impact, funding, taxes or protocol-specific extraction. Performance must then persist across regimes, capital scales and independent or prospective tests. The chain provides a common language for methods that otherwise resist comparison. A time-series foundation model may be valuable because it transfers forecast structure, an LLM because it extracts an event, an RL policy because it internalizes sequential costs, and an agent because it coordinates tools. None receives an exemption from economic validation simply because its architecture is newer or more general.

The same framework also explains why ``financial markets'' cannot be treated as one benchmark. Listed equities and exchange-traded funds (ETFs) operate through scheduled sessions, regulated disclosure, corporate actions, centralized limit-order books, settlement and established shorting and market-access rules. Centralized crypto spot markets trade continuously across fragmented venues, with changing token universes, custody and data-integrity risks. Perpetual futures add funding payments, mark prices, leverage, margin and discontinuous liquidation. On-chain markets expose auditable state and composable transactions but introduce gas auctions, smart-contract risk, automated-market-maker (AMM) inventory loss, transaction reordering and maximal extractable value (MEV). An equity backtest cannot validate a perpetual strategy by analogy; a directionally correct on-chain signal can still lose its surplus to ordering and fees. We consequently keep equities/ETFs, centralized spot, perpetuals and decentralized execution analytically separate, while asking which modeling principles transfer.

Our central conclusion is deliberately bounded. Within the public record examined through 31 August 2026, evidence supports meaningful advances in representation, prediction, textual information processing, portfolio design and operational integration. Selected studies report economically large historical out-of-sample results, including nonlinear equity models and decision-focused policies. The same record does not establish that a general AI method, foundation model or agent architecture delivers persistent, cross-regime and capacity-aware net alpha. The gap is not a semantic technicality. A look-ahead correction can eliminate a celebrated ML alpha \citep{Zhang2025ManMachine}; LLMs can recall pre-training-period financial facts that appear to be forecasts \citep{LopezLira2026Memorization}; cost-unaware learners can load on fleeting signals that sit outside the implementable efficient frontier \citep{Jensen2026ImplementableFrontier}; and prospective AI-managed household portfolios can be concentrated without earning statistically significant abnormal returns \citep{Carlin2026HouseholdAI}. Evidence on AI-labelled hedge funds further suggests that early relative performance can decay as a technology diffuses \citep{Chen2026AIAssetManagement}. These findings do not refute every strategy. They locate where claims become fragile in the evidence examined.

The financial disciplines used here---point-in-time information, explicit benchmarks, implementation costs, capacity and prospective validation---are established principles, not inventions of this review. The contribution is their synthesis and joint application across AI method classes and market strata. First, we connect return prediction, language models, portfolio choice, execution and agents along one investment chain. Second, we define a multidimensional evidence profile and a set of evaluation-design archetypes that distinguish task metrics, historical prediction, portfolio simulation, prospective evaluation, live-capital operation and external persistence without collapsing them into one rank. Third, we apply a common validation framework to stocks and crypto while preserving their different payoff and microstructure equations. Common criteria do not imply equal evidentiary depth: the AI-specific crypto record examined here is thinner than the equity record. Fourth, we turn recurring failure modes into a constructive research agenda: point-in-time data and model checkpoints; economic rather than proxy objectives; joint signal, portfolio and execution evaluation; regime-aware updating; transparent prospective protocols; and controls proportionate to trading authority.

The remaining sections proceed from claims to mechanisms and then back to evidence. Section~\ref{sec:scope} defines the critical-review method and evidentiary boundaries. Section~\ref{sec:translation} formalizes alpha translation and the evidence profile. Sections~\ref{sec:prediction}--\ref{sec:portfolio} assess predictive models, language and agent systems, and portfolio/RL approaches. Section~\ref{sec:markets} compares the four market strata and their execution losses. Section~\ref{sec:evidence} weighs affirmative and contrary profitability evidence. Section~\ref{sec:agenda} develops the research agenda, and Section~\ref{sec:conclusion} states what can and cannot presently be concluded.

%% file: sections/02_scope_method.tex
\section{Scope, Review Method, and Evidentiary Boundaries}
\label{sec:scope}

\subsection{A critical state-of-the-art review, not a systematic review}

This article is a critical state-of-the-art review of research and primary official material available by 31 August 2026. Its purpose is conceptual integration and adversarial evaluation of profitability claims, not exhaustive enumeration or pooled effect estimation. We searched iteratively across finance, economics, machine learning, NLP and market-microstructure venues; followed backward and forward citations around influential primary studies; checked publisher or proceedings records where available; and used first-party documentation for operational interfaces and regulatory controls. We prioritize peer-reviewed articles and official proceedings, then identifiable working papers or dated arXiv versions for rapidly changing topics. Official vendor pages establish that a product or interface existed and what its publisher said it could do; they do not establish return performance. Surveys are used for orientation, not as substitutes for the underlying evidence.

This method has intentional limits. It does not claim database-complete retrieval, duplicate independent screening, a registered protocol, or formal risk-of-bias scores for every record. It is therefore not labelled a systematic review, systematic map or meta-analysis. Quantitative pooling would in any case be misleading for much of this literature: reported Sharpe ratios differ in sampling frequency, annualization, universe, leverage, cash rate, benchmark, risk model, transaction-cost assumptions and model-selection budget. We instead use structured critical comparison. Claims central to the argument are anchored to primary studies, strong contrary findings are retained, and inference is separated from directly reported results. Two ancillary audit aids included with the source package---a claim--evidence map and a representative study-level evidence profile---record the support, contrary evidence and unresolved dimensions behind the main synthesis. They remain selective aids rather than systematic extraction sheets.

\subsection{Inclusion boundaries}

The method boundary is broad but economically specific. We include classical ML, deep learning (DL), time-series foundation models (TSFMs), NLP and LLMs, multimodal models, RL, portfolio and risk optimization, learned execution, and agentic systems when they bear on investment research, allocation or trading. We exclude generic financial chatbots, credit underwriting, fraud detection and corporate forecasting unless a result directly informs traded-asset decisions. ``AI'' is the umbrella term, not a synonym for LLMs. A prompted workflow is not called RL without an explicit sequential learning objective, and tool use is not called autonomy unless the system has documented delegated authority.

The asset boundary covers listed common stocks and ETFs, centralized crypto spot, crypto perpetual futures, and on-chain/DEX markets. It includes cross-asset information when used to trade one of those strata. It does not attempt full treatment of fixed income, foreign exchange, options, commodities, private assets or market making outside the crypto mechanisms needed here. ETFs are included because they are both investable portfolios and an execution vehicle for asset-allocation policies; they are not assumed to be passive. Crypto-linked ETFs remain listed securities for execution analysis even when their underlying exposure is crypto.

\subsection{The unit of evidence}

The unit of evidence is a precisely bounded claim, not a paper-level endorsement. A study can provide strong evidence that a representation predicts returns and weak evidence that the resulting strategy is deployable. We record at least seven dimensions when interpreting an economic result:

\begin{enumerate}[leftmargin=*,itemsep=2pt]
    \item \textbf{Temporality}: Were observations, releases, constituent membership and model artifacts available at the decision timestamp?
    \item \textbf{Selection}: How many features, architectures, prompts, assets, periods and hyperparameters were tried before the reported result?
    \item \textbf{Portfolio mapping}: Are leverage, shorting, concentration, turnover and financing feasible and disclosed?
    \item \textbf{Implementation}: Which commissions, spread, price impact, delay, borrow, funding, gas and failed transactions are included?
    \item \textbf{Risk and benchmark}: Is ``alpha'' measured against an explicit factor model, and are uncertainty and serial dependence handled?
    \item \textbf{External validity}: Does evidence cross assets, venues, regimes, scales or an independently maintained test?
    \item \textbf{Operational provenance}: Is the result a retrospective simulation, a prospectively timestamped paper portfolio, or contemporaneously documented live capital?
\end{enumerate}

Absence of a reported item is coded conceptually as unknown, not favourable. For example, a strategy described as ``net'' is net only of its disclosed cost model. We do not infer zero latency or unlimited borrow because a paper is silent. Likewise, a platform that exposes an order endpoint demonstrates connectivity, not safe operation or investment skill.

\subsection{Reading positive and negative evidence symmetrically}

Financial ML debates are especially vulnerable to asymmetric standards. A large backtest cannot prove permanence, but one failed replication cannot prove universal impossibility. Multiple-testing corrections show why discovery thresholds must rise as researchers search more strategies \citep{Harvey2016FactorZoo,White2000RealityCheck,Bailey2017BacktestOverfitting}; large replication projects then disagree about how much of the published cross section survives. \citet{Hou2020ReplicatingAnomalies} report that a majority of 452 anomalies fail conventional significance in their reconstruction and that an even larger share fails a higher, multiple-testing-aware threshold. In contrast, \citet{Chen2022OpenSource} reproduce almost all of 319 published characteristics under a transparent common code base, and \citet{Jensen2023GlobalFactors} find broad international evidence for many return predictors. The difference reflects definitions, implementations, samples and thresholds as well as genuine empirical disagreement. We preserve it rather than selecting the conclusion that best fits a technological narrative.

The same symmetry applies to trading costs. \citet{NovyMarx2016TransactionCosts} show that costs disproportionately erode high-turnover strategies and that trading rules can change implementability. Using a large proprietary record of institutional trades, \citet{Frazzini2018TradingCosts} report costs lower and capacity higher than several pessimistic estimates for some styles. Neither result licenses a universal cost assumption. A credible AI study should present sensitivity to investor size, participation, urgency and venue, because the cost function is conditional on precisely those variables.

\subsection{What this review means by sustainable net alpha}

We reserve \emph{alpha} for abnormal return relative to a stated risk model, and \emph{net return} for return after explicitly modeled implementation costs. \emph{Sustainable net alpha} is a higher evidentiary standard: positive risk-adjusted performance must survive time-valid external or prospective evaluation across materially different regimes and remain positive at a stated scale. ``Sustainable'' does not mean eternal; it requires a defensible horizon, capacity and monitoring rule. This definition intentionally sets a higher bar than most individual papers attempt. Our conclusion that the public evidence examined here is insufficient at this bar is therefore not the claim that no AI strategy has ever earned money, that proprietary systems cannot work, or that markets are perfectly efficient. It means the examined public record does not support a general, portable promise.

Finally, profitability is not the only legitimate output. AI can reduce research time, standardize evidence retrieval, detect operational exceptions, improve scenario analysis or make a decision trail auditable. Those benefits may be economically important even when incremental alpha is zero. The review nevertheless keeps productivity and investment performance separate: saved analyst hours are not silently added to a Sharpe ratio, and a coherent generated report is not scored as a profitable trade.

%% file: sections/03_alpha_translation.tex
\section{From Model Output to Sustainable Net Alpha}
\label{sec:translation}

\subsection{The alpha-translation chain}

The core analytical object is not a model but a conversion process. Let $\mathcal{I}_t$ denote the point-in-time information set, including the historical version of data and the model available at time $t$. A representation $z_t=f_\theta(\mathcal{I}_t)$ may support a forecast $\hat{r}_{t+1}$, a ranking, an event label or a distribution. A policy $\pi$ converts the representation and current state into target holdings $w_t^*=\pi(z_t,\widetilde{w}_{t-1},c_t)$ under constraints $c_t$, where $\widetilde{w}_{t-1}$ denotes realized post-fill holdings from the preceding decision. An execution rule converts $w_t^*-\widetilde{w}_{t-1}$ into orders; realized fills then determine the post-fill holdings $\widetilde{w}_t$. For a generic portfolio, one-period realized net return can be written schematically as

\begin{equation}
R^{\mathrm{net}}_{p,t+1}=\widetilde{w}_t^{\top} r_{t+1}-C^{\mathrm{fee}}_t-C^{\mathrm{spread}}_t-C^{\mathrm{impact}}_t-C^{\mathrm{delay}}_t-C^{\mathrm{fin}}_t-C^{\mathrm{venue}}_t,
\label{eq:netreturn}
\end{equation}

where every $C_t$ term is a return-normalized deduction for the same measurement period---or, equivalently, a currency cost divided by beginning net asset value---and may depend on submitted orders, realized fills and holdings. Financing includes borrow and leverage costs, and venue-specific losses may include perpetual funding, liquidation, gas, failed transactions or MEV. Taxes are material to many investors but are rarely modeled consistently and are therefore kept explicit rather than presumed. Alpha is the intercept from a stated risk model fitted to $R^{\mathrm{net}}_{p,t}$, not another name for the left-hand side. Sustainable net alpha additionally requires persistence outside the researcher-selected environment and at a stated capital scale.

\begin{figure}[t]
\centering
\begin{tikzpicture}[
  node distance=5mm and 5mm,
  box/.style={draw, rounded corners, align=center, minimum height=9mm, text width=2.35cm, fill=blue!4, font=\small},
  loss/.style={align=center, text width=2.25cm, font=\scriptsize, text=red!65!black},
  monitor/.style={draw, dashed, rounded corners, align=center, inner xsep=4mm, inner ysep=2mm, fill=gray!5, font=\scriptsize, text=gray!75!black},
  arr/.style={-{Latex[length=2mm]}, thick},
  feedback/.style={-{Latex[length=2mm]}, dashed, gray!75}
]
\node[box] (data) {Point-in-time\\information};
\node[box, right=of data] (signal) {Representation\\and signal};
\node[box, right=of signal] (policy) {Decision, portfolio\\and risk};
\node[box, right=of policy] (exec) {Orders, fills\\and costs};
\node[box, right=of exec] (evid) {Net alpha, capacity\\and persistence};
\draw[arr] (data) -- (signal);
\draw[arr] (signal) -- (policy);
\draw[arr] (policy) -- (exec);
\draw[arr] (exec) -- (evid);
\node[loss, below=5mm of data] {revision, survivorship, model-memory leakage};
\node[loss, below=5mm of signal] {overfit, weak labels, unstable regimes};
\node[loss, below=5mm of policy] {turnover, leverage, concentration, borrow};
\node[loss, below=5mm of exec] {spread, impact, latency, funding, gas, MEV};
\node[loss, below=5mm of evid] {selection, crowding, scale, governance failure};
\node[monitor, above=9mm of policy] (monitor) {Prospective monitoring\\controlled updating};
\draw[feedback] (evid.north) |- (monitor.east);
\draw[feedback] (monitor.west) -| (data.north);
\end{tikzpicture}
\caption{The alpha-translation chain. Technical improvement at one stage is economically valuable only to the extent that information survives all downstream transformations. Red labels identify representative failure channels, not an exhaustive list.}
\label{fig:translation}
\end{figure}
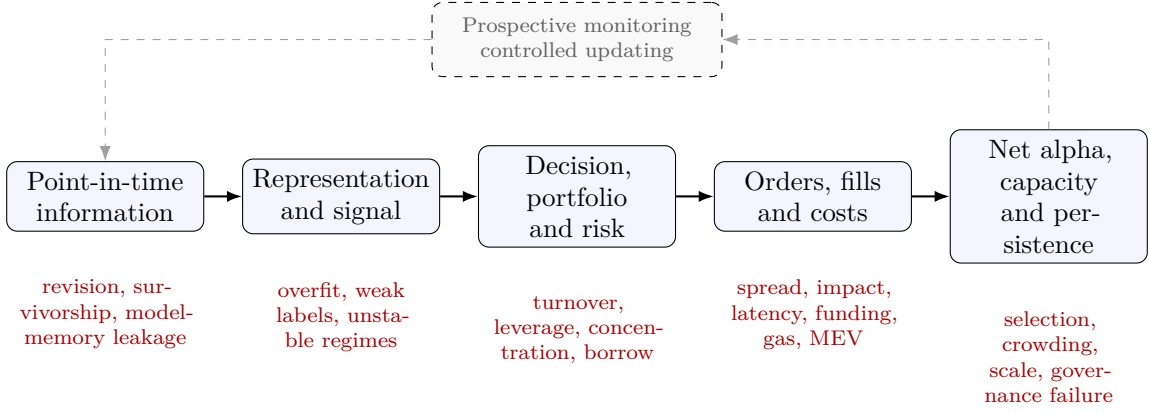

Figure~\ref{fig:translation} shows why evidence can weaken even while components improve. Each arrow is a possible loss of validity. Revised fundamentals, delisted firms or pretraining memories can contaminate the information set. A high-$R^2$ forecast can be concentrated in low-capacity securities. A ranker can induce excessive turnover. A target portfolio can require unavailable borrow. A correct order can be filled after its signal decays. A profitable simulation can disappear when many investors deploy it, or it can be operationally unacceptable because no one can reconstruct the source, model version or authorization path. The last box feeds back to earlier stages because performance and controls must be monitored rather than frozen after deployment.

\subsection{A multidimensional evidence profile, not a leaderboard}

Evidence quality is a profile rather than a scalar tier. For each economic claim, we assess the seven dimensions introduced in Section~\ref{sec:scope}: temporality ($T$), selection control ($S$), portfolio mapping ($P$), implementation realism ($I$), risk and benchmark specification ($R$), external validity ($X$), and operational provenance ($O$). The profile is

\begin{equation}
\mathcal{E}=(T,S,P,I,R,X,O).
\label{eq:evidenceprofile}
\end{equation}

These dimensions are not interchangeable. A live record can have strong operational provenance but weak selection control; a carefully designed historical study can isolate a mechanism while saying little about capacity or future persistence. Table~\ref{tab:evidence_profile} therefore lists recurring evaluation-design archetypes, not ranked levels. The codes identify the dominant evidence object and do not imply a total ordering. A task benchmark becomes evidence about net, prospective or live profitability only when the corresponding dimensions are separately documented.

\begin{table}[t]
\centering
\caption{Evaluation-design archetypes within a multidimensional profitability-evidence profile. Rows are common study designs, not a total ordering; each must be assessed on all seven dimensions in Equation~\ref{eq:evidenceprofile}.}
\label{tab:evidence_profile}
\scriptsize
\begin{tabularx}{\textwidth}{@{}p{1.15cm}p{2.75cm}Y Y@{}}
\toprule
Code & Evidence object & Primary support & Dimensions unresolved unless reported \\
\midrule
CAP & Architecture or vendor capability & Feasibility of representation, tool use or order connectivity & Prediction, portfolio mapping and return \\
TASK & Task metric & Classification, extraction, reasoning or forecast accuracy on a defined benchmark & Feasible positions, implementation and economic value \\
OOS & Time-valid statistical test & Incremental prediction after a declared historical split & Portfolio value, costs and capacity \\
GROSS & Gross portfolio simulation & Historical decision value under stated weights and constraints & Implementable net performance \\
NET & Net, risk-adjusted simulation & Simulated alpha under disclosed cost, financing and execution models & Prospective robustness and unmodeled costs \\
PROSP & Precommitted paper or shadow portfolio & Resistance to hindsight and changing model selection & Real fills, operational loss and live-capital behavior \\
LIVE & Audited live-capital record & Orders, fills, costs and controls for the observed size and period & Causality, future persistence and larger capacity \\
EXT & Independent cross-regime/capacity study & A bounded sustainability claim for specified settings & A guarantee under future competition or structural change \\
\bottomrule
\end{tabularx}
\end{table}

Three distinctions deserve emphasis. First, ``out of sample'' is not synonymous with point-in-time. A test observation can be held out while the input database contains later revisions, the universe contains future survivors, or the pretrained LLM remembers events from its corpus. The appropriate condition is

\begin{equation}
\mathcal{I}_t \cap \mathcal{F}_{t+1}=\varnothing,
\end{equation}

where $\mathcal{F}_{t+1}$ denotes information observable only after the decision. This must hold for feature values, timestamps, membership, preprocessing estimates, prompt examples and model parameters. The correction documented by \citet{Zhang2025ManMachine} illustrates the stakes: a reported ML return advantage can vanish when predictors are aligned to their actual availability.

Second, a cost deduction is a model, not a stamp of realism. Linear costs may approximate commissions and spread at small size; impact is nonlinear and depends on volatility, liquidity, urgency and participation. If target holdings are $q_t$ and the strategy trades $\Delta q_t$, then writing $C_t=k|\Delta q_t|$ silently assumes away much of the capacity problem. \citet{Jensen2026ImplementableFrontier} show that cost-unaware return predictors can favor fleeting, low-scale opportunities and produce unattractive implementable frontiers, whereas learning the portfolio with trading costs inside the objective changes which information is valuable. The result is more than ``subtract a higher fee.'' Economic constraints alter the optimal representation and decision rule.

Third, a statistically significant alpha is not necessarily decision-relevant. Investors care about expected utility, drawdown, tail exposure, funding stability and the covariance of a strategy with the rest of their wealth. Mean--variance choice remains a useful abstraction \citep{Markowitz1952Portfolio}, but estimated means are noisy and optimized portfolios are unstable. The long-standing difficulty of beating simple $1/N$ diversification out of sample \citep{DeMiguel2009OptimalVersusNaive} is a warning against adding an elaborate optimization layer without measuring estimation error. AI can reduce some approximation error while increasing selection error and opacity.

\subsection{A decomposition of the central verdict}

The central verdict follows from a mismatch between the location of progress and the location of decisive evidence in the reviewed record. Most documented public advances are upstream: representation learning, scalable prediction, text extraction, workflow integration and historical policy optimization. The most demanding profitability claims are downstream: timestamped commitment, real execution, complete costs, meaningful risk adjustment, capacity and persistence. This is why the review can affirm material changes in the investment practices examined while concluding that the reviewed public evidence does not establish a general profit engine.

The verdict is also conditional on disclosure. Proprietary firms may possess better data, engineering and live evidence than they publish. Public insufficiency is not evidence of private absence. But unpublished performance cannot validate public architectural claims, and selective disclosure creates its own inference problem. A scientific claim must be supported by inspectable evidence at the level claimed. The practical response is not to demand that every study operate a fund. It is to report the full evidence profile and design the next evaluation around the dimensions that remain unresolved.

%% file: sections/04_prediction_representation.tex
\section{Prediction and Representation: Real Progress, Conditional Economic Value}
\label{sec:prediction}

\subsection{From linear characteristics to nonlinear cross-sectional models}

The strongest case that ML has changed empirical investing comes from large, structured equity panels. Traditional predictive regressions impose a small set of linear effects because expected returns are noisy and the number of candidate interactions grows rapidly. Regularization, trees, boosting and neural networks relax this restriction while controlling complexity. In a broad US sample, \citet{Gu2020EmpiricalAssetPricing} compare multiple ML methods using firm characteristics and macroeconomic variables. Trees and neural networks improve out-of-sample prediction and portfolio performance relative to standard linear benchmarks; the result also shows that relatively shallow architectures can be better suited to financial data's low signal-to-noise ratio than indiscriminately deeper networks. \citet{Chen2024DeepAssetPricing} use a deep stochastic-discount-factor framework that joins no-arbitrage restrictions, macroeconomic state and firm characteristics. Here, economic structure is not decoration after prediction: it shapes the learning problem.

These studies establish that nonlinearities and high-dimensional interactions contain information under their protocols. They do not establish universal alpha. Their historical universes, rebalancing rules and institutional assumptions define a particular opportunity set. Cross-sectional predictability may compensate investors for risk, reflect mispricing, or combine both, and a factor-model intercept depends on which risks the model recognizes. Moreover, portfolio sorts amplify small statistical differences and can concentrate exposure in illiquid stocks. In the profile of Table~\ref{tab:evidence_profile}, their support is strongest on time-valid historical prediction and disclosed portfolio simulations; prospective, capacity and live-operation dimensions remain open.

Replication evidence further resists a simple story. The multiple-testing analysis of \citet{Harvey2016FactorZoo} implies that conventional significance thresholds are too permissive when researchers search hundreds of candidate factors. \citet{Hou2020ReplicatingAnomalies} find that many published anomalies fail to replicate under their procedures. Yet \citet{Chen2022OpenSource} reproduce most signals in a transparent, standardized library, while \citet{Jensen2023GlobalFactors} document a broad global cross section. The lesson for AI is not that characteristics are either ``real'' or ``dead.'' Feature definition, accounting lag, microcap treatment, weighting, exchanges, sample end and statistical threshold are part of the model. A neural network trained on a disputed panel can fit the disagreement rather than resolve it.

The time dimension adds another challenge. \citet{WelchGoyal2008EquityPremium} show how equity-premium predictors that fit in sample can fail out of sample, and \citet{Goyal2024Comprehensive} document substantial deterioration in a long update of predictive variables. \citet{McLeanPontiff2016Publication} report lower returns after sample end and especially after academic publication, consistent with both statistical overfit and trading that attenuates mispricing. Thus a predictive relationship can be genuine in the discovery sample and still be a poor forecast of future strategy profitability. AI expands the hypothesis space and accelerates discovery, which increases both the chance of finding useful nonlinear structure and the burden of controlling selection.

\subsection{Deep learning and economically structured policies}

DL contributes in three different ways that should not be conflated. Representation models compress high-dimensional prices, characteristics or text; prediction models estimate conditional returns, risks or state transitions; decision models output positions or execution actions. Most reported gains occur in one of these roles, and the downstream evidence depends on the others.

The promise of deep representation is greatest where interactions are numerous and economically constrained data are sufficiently broad. Asset-pricing networks can enforce a pricing equation, attend across firms or learn latent states instead of treating observations as independent rows. But financial depth has a different meaning from depth in computer vision: samples are correlated, regimes change, and an additional decade provides only a small number of independent crises. Model size is not an adequate proxy for effective sample size. Large parameter counts can be defensible when pretraining transfers stable structure or when cross-sectional breadth is informative; they can also create a large, weakly observed search space.

Decision-focused work addresses a more consequential limitation: minimizing forecast error need not maximize investor welfare. \citet{Simon2026DeepPolicies} estimate parametric portfolio policies that map characteristics to holdings under an investor objective, reporting higher historical out-of-sample certainty-equivalent returns than linear policies across their setting and robustness exercises. \citet{Jensen2026ImplementableFrontier} formulate ML around the implementable efficient frontier. Their comparison is revealing: standard prediction-oriented ML extracts signals that can appear valuable gross but trade poorly after costs, particularly when opportunities are fleeting and low scale; directly cost-aware learning changes both portfolio composition and the attainable frontier. These results support a design principle---train for the decision actually made---rather than a claim that one neural architecture dominates all markets.

Direct objectives introduce their own specification risk. A policy trained to maximize an estimated Sharpe ratio can exploit volatility estimates, boundary constraints or leverage conventions. A drawdown penalty encodes a horizon and path dependence that may not match investor liabilities. Transaction costs calibrated from a representative account can be wrong for another scale. Economic training therefore moves assumptions inside the loss function; it does not eliminate them. Robust evaluation must perturb risk aversion, financing, cost curves, turnover limits, short availability and execution delay, and it should report which performance survives rather than selecting the most favourable preference.

\subsection{Time-series foundation models}

TSFMs extend pretraining and zero-shot transfer to numerical sequences. TimesFM is pretrained on a large time-series corpus and reports competitive zero-shot forecasting across general benchmarks \citep{Das2024TimesFM}. Moirai is trained on the LOTSA collection, spanning many domains, frequencies and variable dimensions, and is evaluated as a universal forecaster \citep{Woo2024Moirai}. Chronos tokenizes numerical values and uses language-model-style training for probabilistic forecasting \citep{Ansari2024Chronos}. Time-MoE scales a mixture-of-experts time-series model and reports strong general forecasting performance \citep{Shi2025TimeMoE}. These are substantial engineering and statistical advances: a common pretrained model can provide usable forecasts without fitting a separate architecture from scratch for every dataset.

Their direct profitability evidence is much weaker. General time-series benchmarks are designed around forecasting loss, not tradable excess return. Asset prices are close to nonstationary, economic value often lies in a small conditional mean relative to volatility, and a lower mean absolute error can result from better level or seasonal prediction without improving the sign, timing or capacity of a trade. A universal model can forecast realized volatility well and still require a separate risk-allocation thesis. Conversely, it may be economically useful even when directional-return accuracy is unchanged, because better uncertainty or covariance forecasts can improve sizing. The model's role must be stated before its benchmark metric is interpreted.

Two finance-specific preprints available by the cutoff sharpen this mixed result without turning forecast accuracy into trading profit. In rolling-origin tests on five liquid US equities, \citet{Alonso2026TSFMReturns} find that pretrained TSFMs take eight of ten task-level wins, yet improvements over a random-walk benchmark are small and sparse, with statistically significant gains in only two model--asset comparisons. FinVerse broadens the measurement substrate to 116,897 financial series and compares 43 public foundation models across domain-aligned metrics; its authors find that strong generic forecasting ranks need not translate into useful financial forecasts \citep{Lee2026FinVerse}. Both are author-run preprints. Neither supplies a portfolio with execution costs, capacity, or live-capital returns.

There is also a caution against assuming that a large pretrained language backbone is intrinsically useful for numerical forecasting. \citet{Tan2024AreLLMsUseful} systematically ablate several LLM-based time-series methods and find that removing or replacing the LLM component can preserve or improve performance, while claimed few-shot benefits do not reliably appear in their tests. This is not a refutation of time-series pretraining: TimesFM, Moirai and Chronos are trained around numerical sequences rather than relying on semantic language knowledge. It is evidence that transfer must be demonstrated against smaller, time-series-specific and non-pretrained baselines under equal information and tuning budgets.

Financial application raises a further point-in-time issue. A generic TSFM may have been trained on historical series extending into the intended test period. Unlike a text model that can sometimes disclose a corpus date, numerical pretraining sets may contain transformed or duplicated benchmark sequences whose provenance is difficult to audit. A defensible investment experiment needs a model checkpoint whose entire training data ends before the trading period, or a contamination test strong enough to bound the overlap. Fine-tuning only on a pre-cutoff slice does not remove information already encoded in the base model.

\subsection{What predictive metrics reveal---and conceal}

Table~\ref{tab:method_evidence} summarizes the main method families by their demonstrated contribution and their characteristic translation failure. It is not a ranking: studies inside each family vary widely, and the table records the strongest defensible generalization rather than the largest reported return.

\begin{table}[t]
\centering
\caption{Method families viewed through the alpha-translation chain. ``Progress'' denotes an evidence-backed technical contribution, not a general profitability finding.}
\label{tab:method_evidence}
\small
\begin{tabularx}{\textwidth}{@{}p{2.25cm}Y Y Y@{}}
\toprule
Family & Demonstrated progress & Principal economic opportunity & Characteristic evidence gap \\
\midrule
Regularized ML, trees, boosting & Nonlinear and high-dimensional return prediction in large panels & Ranking and interaction discovery & Multiple testing, unstable features, liquidity concentration \\
Deep asset-pricing models & Joint representation and economic restrictions & Latent states, nonlinear stochastic discount factors & Effective sample size, interpretation, historical dependence \\
TSFMs & Zero-shot and cross-domain forecast transfer & Fast adaptation, risk and covariance inputs & Generic loss is not alpha; pretraining contamination \\
Financial NLP/LLMs & Scalable extraction and semantic event processing & Timely news, filing and expectation signals & Memorization, timestamping, prompt selection, source provenance \\
RL and direct policies & Sequential or decision-aligned objectives & Turnover, constraints and dynamic risk inside learning & Simulator error, reward hacking, fragile policy transfer \\
LLM agents & Integration of retrieval, memory, reasoning and tools & End-to-end research and operational coordination & Component attribution, reproducibility, cost and live validation \\
\bottomrule
\end{tabularx}
\end{table}

Predictive $R^2$, classification accuracy and information coefficients are useful diagnostic quantities. They reveal whether a model adds conditional information and where. They conceal the portfolio's nonlinear exposure to tails, the covariance among simultaneous signals, trading required to realize a forecast, and the value of abstention. A small, stable rank correlation in liquid securities can be more valuable than high accuracy in rare or untradeable cases. An apparent directional hit rate can be produced by market drift. Calibration matters when position size depends on confidence; ranking may matter when only a fixed long--short book is feasible. No single predictive metric dominates without an explicit decision problem.

The best reading of this literature is therefore two-sided. It is no longer credible to claim that only linear models can extract useful information from financial data. It is equally unsupported to infer from upstream benchmark gains that general pretrained models have solved investing. Prediction is a valuable and increasingly transferable component. Its economic value is conditional on temporal provenance, stable incremental information and a downstream policy that survives risk and execution.

%% file: sections/05_text_multimodal_agents.tex
\section{Text, Multimodality, and Agents as an Integration Layer}
\label{sec:agents}

\subsection{Text can contain return-relevant information}

Financial text is an unusually plausible domain for AI because prices respond to information that is not naturally represented as a fixed numerical panel. Early work established both the opportunity and the domain problem. \citet{Tetlock2007Media} links pessimistic media content to market activity and subsequent return dynamics. \citet{Loughran2011Liability} shows that general-purpose sentiment dictionaries misclassify financially meaningful language, motivating domain-specific vocabularies. Supervised methods can learn a return-predictive text representation rather than treating sentiment as a universal label; \citet{Ke2019TextReturns}, for example, construct a text-based signal aligned with return prediction. These studies precede generative LLMs but remain conceptually important: the economic label, release time and audience expectation matter more than fluent language generation.

Modern financial language models expand scale and task coverage. BloombergGPT demonstrates finance-intensive pretraining across financial and general corpora \citep{Wu2023BloombergGPT}; FinGPT proposes an open-source ecosystem for adapting language models to financial data and tasks \citep{Yang2023FinGPT}. Fine-tuned models classify news, extract entities and relations, summarize filings, answer questions and generate research. Multimodal systems can add price patterns, tables, charts or audio transcripts. These capabilities lower the cost of converting heterogeneous disclosure into structured features. They can also improve analyst productivity without making any trade, a benefit that should be measured through time saved, coverage, error rates and decision quality rather than recast as alpha.

The strongest recent profitability-related LLM evidence comes from designs that respect event time. Using news headlines after the model's training-data cutoff, \citet{LopezLira2026ChatGPT} find that GPT classifications are informative about subsequent stock-price movements, especially around smaller firms and negative news. The design helps separate semantic processing from historical recall, and the paper documents that predictability weakens as use of the technology spreads. The interpretation requires care: classifying the market's immediate reaction is not a trade available before that reaction, whereas subsequent drift is potentially actionable subject to release timestamps, latency, costs and capacity. The result is evidence that an LLM can extract economically relevant event information, not that an off-the-shelf chat interface supplies a permanent strategy.

Two lines of work clarify temporal validity. \citet{LopezLira2026Memorization} show that LLMs can recall financial values and events from their pretraining period, and that masking obvious identifiers is not a sufficient defence. An apparently prescient historical narrative can therefore be retrieval from parameters. \citet{Kelly2026PITLanguageModels} train a sequence of point-in-time language-model checkpoints with information cutoffs, enabling historical tests without using a model trained on later text. Their evidence suggests that point-in-time representations retain useful financial information and can approach the apparent performance of contaminated alternatives in selected tests. The broader lesson is constructive: contamination is not a reason to abandon LLM research, but a reason to version the base model, retrieval corpus, prompt, tool data and evaluation period as a single time-indexed object.

\subsection{Why multimodality helps and complicates identification}

An investor rarely observes one modality in isolation. Earnings news is interpreted relative to prior guidance, estimates, prices, options and macro conditions; an on-chain transfer may mean something different depending on exchange flows and perpetual positioning. Multimodal models can represent these interactions without forcing every source into a handcrafted scalar. FinAgent combines textual and market information with tools in a generalist trading agent \citep{Zhang2024FinAgent}; CryptoTrade combines on-chain and off-chain evidence with reflection for cryptocurrency decisions \citep{Li2024CryptoTrade}. Such systems illustrate a useful direction: information fusion is aligned with the actual analyst problem.

Yet adding modalities increases the number of timestamps, revisions and missingness mechanisms that must be correct. A news database may record ingestion rather than first public release. Fundamental values may be backfilled. A chart image can encode future axes or annotations. Blockchain state is timestamped but may not be final at the moment a transaction is signed. If a multimodal ablation removes an entire feed, it may reveal that the feed matters but not whether the model fused it intelligently; a simpler late-fusion baseline or decision rule may capture the same value. Comparisons should hold the information set, latency and search budget fixed before attributing performance to architectural integration.

\subsection{Agentic systems: architectural closure before evidentiary closure}

LLM agents shift the unit of design from a single prediction to a repeated workflow. A typical system retrieves news and fundamentals, maintains episodic or hierarchical memory, assigns specialized analyst roles, debates opposing theses, calls quantitative tools, generates a position and revisits the decision after outcomes. FinMem emphasizes layered memory and character design \citep{Yu2023FinMem}; FinCon uses multiple agents and verbal reinforcement \citep{Yu2024FinCon}; TradingAgents organizes analyst, researcher, trader, risk and portfolio roles in a multi-agent framework \citep{Xiao2024TradingAgents}; and FinAgent combines multimodal observations and tool use \citep{Zhang2024FinAgent}. These projects are important because they make the research-to-action boundary explicit. They are also difficult to evaluate: an end-to-end return mixes information sources, base-model knowledge, prompts, role decomposition, tools, risk rules and random generation.

An agent is therefore an integration layer, not a new asset-pricing law. Debate can reduce one-sided reasoning, but correlated agents using the same base model may create verbose agreement rather than independent evidence. Memory can preserve a thesis, but can also anchor the system to a past regime. Reflection can update behavior, but evaluating on the same outcomes used for reflection converts the test into training. Tool use can improve arithmetic and access fresh data; it can also enlarge the prompt-injection, credential and erroneous-order surface. The appropriate baselines include a single agent with the same model and tools, deterministic rules using the same signals, equal token and data budgets, and ablations that preserve timestamps.

Benchmark design is beginning to strengthen the temporality and operational-provenance dimensions of the evidence profile. StockBench emphasizes real-world stock trading evaluation across a temporally structured setting \citep{Chen2025StockBench}; Agent Trading Arena studies numerical understanding and trading behavior in a controlled arena \citep{Ma2025AgentTradingArena}; DeepFund commits fund recommendations in real time to prevent retrospective ``time travel'' \citep{Li2025DeepFund}; and Agent Market Arena evaluates agents in a live, multi-market setting \citep{Qian2026AgentMarketArena}. These efforts improve prospective validity and comparability. A live feed or paper arena still differs from audited live-capital execution, however, and short evaluation windows can be dominated by market beta, style exposures or a few events. The research frontier is less about adding roles than about longer timestamped evaluation, complete holdings and fill logs, predeclared model updates, and uncertainty that reflects both market and agent stochasticity.

A contemporaneous agentic-quantitative-trading survey reaches a compatible diagnosis: current systems remain concentrated on signal discovery, while complete integration with portfolio construction, execution and risk control is uncommon \citep{Hua2026AgenticQuantTrading}. A separate audit-oriented preprint maps 77 studies but identifies only 19 with both action output and closed-loop evaluation; within that subset it reports two time-consistent splits, one explicit transaction-cost model, one documented universe or survivorship treatment, and no fully replayable R3 record \citep{Xia2026AgenticTrading}. These author-coded counts diagnose protocol and reporting gaps, not the profitability of every agentic system. Our broader market-wide synthesis uses both surveys as context rather than as return evidence.

Table~\ref{tab:agent_systems} locates representative systems and official interfaces by what they demonstrate. Academic papers are evidence for their reported experiment; official product documentation is evidence that a capability is offered. Neither category is silently upgraded to durable profitability evidence.

\begin{table}[t]
\centering
\caption{Representative agentic integration systems and interfaces. The table describes the principal evidenced contribution, not a performance ranking.}
\label{tab:agent_systems}
\scriptsize
\begin{tabularx}{\textwidth}{@{}p{2.2cm}p{1.55cm}Y Y@{}}
\toprule
System/source & Market emphasis & Integration contribution & Profitability boundary \\
\midrule
FinMem \citep{Yu2023FinMem} & Equities & Layered memory and reflective trading workflow & Retrospective experiments; memory benefit is conditional on protocol \\
FinAgent \citep{Zhang2024FinAgent} & Equities & Multimodal observations, tools and generalist agent & Historical evaluation, not independent live-capital persistence \\
FinCon \citep{Yu2024FinCon} & Equities/\allowbreak other tasks & Multi-agent roles and conceptual verbal reinforcement & Reported backtests do not isolate a general alpha mechanism \\
TradingAgents \citep{Xiao2024TradingAgents} & Selected US equities & Analyst debate, trader and risk/portfolio roles & Short historical test on selected stocks; architecture is not durable-alpha proof \\
StockBench \citep{Chen2025StockBench} & Equities & Time-structured benchmark for LLM-agent trading & Benchmark outcome remains model-, window- and cost-specific \\
Agent Trading/Market Arenas \citep{Ma2025AgentTradingArena,Qian2026AgentMarketArena} & Controlled and multi-market & Comparative numerical behavior and live-market evaluation & Prospective arenas improve validity but need longer, fill-aware, risk-adjusted records \\
CryptoTrade \citep{Li2024CryptoTrade} & Crypto spot & On-chain/off-chain fusion and reflection & Beats selected learned baselines; does not establish superiority to all traditional signals \\
DSA \citep{Zhu2026DSA} & Multi-market stocks & Evidence-aware research orchestration and report support & Research/report generation; conformance evidence only, with no report-quality, forecast, return, or execution result \\
Robinhood, Alpaca, Coinbase \citep{Robinhood2026Agentic,Alpaca2026MCP,Coinbase2024AgentKit} & Brokerage and crypto execution & Agent-facing research, account and order/wallet tools with stated controls & First-party capability claims; no independent evidence of alpha \\
\bottomrule
\end{tabularx}
\end{table}

\subsection{Brokerage and on-chain tools change the risk boundary}

The release of agent-facing trading interfaces makes the distinction between reasoning and authority operationally urgent. Robinhood's official material distinguishes the earlier Cortex research product from an Agentic Trading account whose documented interface can expose account data and order placement under account-specific permissions \citep{Robinhood2025Cortex,Robinhood2026Agentic,Robinhood2026Support}. In a July 2026 SEC exhibit, Robinhood reported nearly 100,000 opened Agentic Trading accounts and more than USD~100 million in assets under custody \citep{Robinhood2026Q2}. Those are company-reported adoption figures, not counts of active or profitable accounts, evidence of safe use, or an independent audit. Alpaca's official Model Context Protocol server exposes market and account operations, including order-related endpoints; its documentation recommends paper testing and warns of unintended orders, slippage and limits \citep{Alpaca2025MCP,Alpaca2026MCP}. Coinbase AgentKit provides wallet and blockchain actions for agents \citep{Coinbase2024AgentKit}. These products lower integration cost. They do not show that an agent selects good trades.

Tool access, delegated authority and investment skill are separate variables. A read-only research agent may create reputational or confidentiality risk but cannot place an order. An order-capable agent can create immediate market and credit exposure even when its forecast is unchanged. Controls should therefore be attached to authority: allow-listed instruments and venues, maximum notional and leverage, price collars, stale-data rejection, rate limits, transaction simulation, human approval above thresholds, independent position reconciliation and a kill switch outside the model's control. Smart-contract calls additionally require chain, contract and token allow-lists, allowance limits, gas bounds and protection against malicious tool output. The successful demonstration of a buy order is CAP-type operational evidence and a governance event, not a profitability experiment.

Agents may ultimately be most valuable as disciplined coordinators: they can keep evidence and counterarguments visible, call deterministic calculators, enforce a checklist, and document why a decision changed. That value is credible only if provenance survives generation. Every consequential statement should link to a dated source or computed object; each tool result should carry a timestamp; and the final action should be reproducible from logged inputs, versions and policy constraints. Fluent rationales generated after an action are not provenance. The frontier is thus an auditable hybrid: learned components for extraction and adaptation, deterministic components for accounting and limits, and explicit human or institutional responsibility for delegated risk.

%% file: sections/06_portfolio_rl_risk.tex
\section{Portfolio Learning, Reinforcement Learning, and Risk}
\label{sec:portfolio}

\subsection{The portfolio is where weak assumptions become positions}

Prediction papers often postpone the portfolio problem by forming equal-weighted quantiles. That convention is transparent, but a deployed investor must choose exposure, covariance risk, turnover, financing and abstention jointly. Mean--variance optimization formalizes the trade-off between expected return and variance \citep{Markowitz1952Portfolio}; in practice, expected returns are estimated so imprecisely that unconstrained optimized weights can be extreme. The out-of-sample strength of simple diversification documented by \citet{DeMiguel2009OptimalVersusNaive} remains a relevant baseline for AI. A complicated allocator should be compared not only with another learned model but with equal weight, volatility scaling, a market portfolio, a low-turnover characteristic strategy and a no-trade policy.

End-to-end portfolio learning attempts to optimize what investors value rather than the error of an intermediate forecast. Direct policies can learn nonlinear mappings from features to weights, and risk aversion can regularize those mappings economically \citep{Simon2026DeepPolicies}. AlphaPortfolio uses a deep RL formulation for dynamic portfolio optimization and reports large historical out-of-sample risk-adjusted performance under its protocol \citep{Cong2026AlphaPortfolio}. Such results are affirmative evidence that economic losses and sequential decisions can change the frontier. They remain historical evidence: the strategy's attainable return depends on the chosen asset universe, action interval, transaction-cost curve, state variables and constraints, as well as whether the research search that produced the architecture is counted in uncertainty.

Cost-aware portfolio learning is especially important because signal value depends on how slowly it decays. A small persistent signal can support a large, patient position; a stronger one-step forecast may be consumed by spread and impact. \citet{Jensen2026ImplementableFrontier} make this interaction explicit by learning the implementable efficient frontier rather than attaching costs to a preselected forecast. Related portfolio-combination work shows that combining predictors can cancel offsetting trades and improve implementability \citep{DeMiguel2020TransactionCostManaged}. The general principle is that turnover is not merely an evaluation statistic. It is a state variable and design constraint.

\subsection{What RL adds}

RL is appropriate when current actions change future states or opportunities: inventory affects later execution, trading changes transaction costs and tax lots, leverage changes liquidation risk, and an order can reveal information. A policy $\pi(a_t\mid s_t)$ is trained to maximize expected discounted or finite-horizon reward,

\begin{equation}
J(\pi)=\mathbb{E}_{\pi}\left[\sum_{t=0}^{T}\gamma^t u\!\left(R^{\mathrm{net}}_{p,t},x_t\right)\right],
\end{equation}

where $x_t$ may include drawdown, inventory or funding state. Early recurrent reinforcement learning directly optimized a risk-adjusted trading objective \citep{Moody2001RRL}. Later work applies deep RL to allocation and execution; FinRL-Meta and TradeMaster package datasets, environments, baselines and evaluation tools intended to improve reproducibility \citep{Liu2022FinRLMeta,Wang2023TradeMaster}. These infrastructures are meaningful progress because inconsistent data preparation and environments can dominate algorithm comparisons.

RL also magnifies the simulation problem. Historical markets do not reveal the counterfactual prices and queues that would have followed a different large action. A backtest environment commonly assumes that the agent is a price taker, fills at the next bar, and observes a Markov state assembled from historical variables. Under those assumptions an algorithm can learn simulator shortcuts: trading at a bar price unavailable in real time, exploiting a reward normalization fitted with future data, or using turnover patterns that would alter impact. A policy's repeated interaction with a fixed historical tape is not interactive market evidence.

Crypto provides instructive negative as well as positive results. The recurrent RL crypto agent of \citet{Borrageiro2022CryptoAgent} reports a systematic framework for crypto trading, whereas a PPO study of BTC/USDT minute data reports failure to outperform buy-and-hold under its experiment \citep{Ferreira2022CryptoDRL}. The contrast is healthier than a leaderboard: performance depends on period, benchmark, action space, reward and cost model. Buy-and-hold is itself a high-beta, regime-dependent comparator, so beating or failing to beat it does not identify alpha. Crypto RL should additionally compare volatility-matched exposure, cash, momentum or carry baselines and state whether funding and liquidation are modeled.

\subsection{Risk is not a penalty appended at the end}

Many AI systems represent risk as one scalar penalty. Real portfolio risk is multidimensional: market and factor exposure, concentration, correlation, drawdown, liquidity, short squeeze, borrow recall, counterparty, custody, stablecoin, smart-contract and operational risk. A variance penalty treats upside and downside symmetrically and can miss liquidation discontinuities. A drawdown constraint is path-dependent. Value-at-risk or expected shortfall depends on tail estimation precisely where data are sparse and regimes change. Risk models should therefore be stress-tested as models, not treated as ground truth.

Risk also determines whether a return is interpretable. A long-only agent evaluated during a bull market can earn positive return through beta while adding no security-selection value. A crypto policy can earn a high nominal return by loading on market direction, small tokens or short-volatility carry. Appropriate reporting includes gross and net return, volatility, drawdown, turnover, leverage, average and peak concentration, beta and factor alpha where meaningful, tail loss, and exposure to the venue-specific cash flows discussed in Section~\ref{sec:markets}. Confidence intervals should respect time dependence; repeated runs are required when generated decisions or RL training are stochastic.

Distribution shift is not one generic robustness test. A policy can face covariate shift in features, label shift in return distribution, a new market mechanism, missing vendors, a changed fee tier, an asset delisting or an adversarial text source. Training with historical perturbations may improve resilience but can also create false comfort if stress scenarios omit the actual discontinuity. A robust deployment separates monitored assumptions from learned flexibility: hard limits remain outside the policy; state drift and performance attribution trigger review; model updates are staged through shadow operation; and rollback does not require the failing model to cooperate.

\subsection{Execution as a sequential decision problem}

Execution is one domain where sequential learning has a natural target: buy or sell a required quantity while balancing price risk and market impact. Classical optimal execution already formalizes this trade-off \citep{AlmgrenChriss2001Execution}, and market-impact theory explains why private information and order flow affect prices \citep{Kyle1985ContinuousAuctions}. RL can adapt order placement to book state and partial fills; \citet{Nevmyvaka2006Execution} provide an early empirical application to optimized trade execution. The relevant benchmark is an execution schedule with the same parent order and information, not the portfolio return created by the upstream signal.

Execution evaluation should report implementation shortfall against a predeclared arrival or decision price, fill probability, adverse selection, latency, rejected orders and tail outcomes. Training and testing on messages from one venue can overfit queue priority and fee rules. A policy that improves average shortfall but occasionally fails to complete a risk-reducing order may be unacceptable. Thus even in a domain where RL's sequential advantage is credible, optimization requires hard completion, credit and price controls.

The synthesis is that portfolio and RL methods address real weaknesses of forecast-first investing. They can internalize constraints, dynamic costs and risk, and strong historical results justify further research. They also relocate model risk into the objective, simulator and constraint set. A decision-aligned loss is better than a proxy only when the decision environment is time-valid and economically faithful.

%% file: sections/07_market_strata_execution.tex
\section{Market Structure and the Meaning of Profitable Prediction}
\label{sec:markets}

\subsection{Listed equities and ETFs}

Equities provide the deepest public record for ML asset pricing: long return histories, standardized accounting variables, regulated corporate disclosures and established factor benchmarks. They also contain traps that a model can exploit. Delisted firms and delisting returns must be represented to avoid survivorship bias \citep{Shumway1997DelistingBias}; accounting and index data must be lagged to publication; corporate actions alter prices and shares; and historical constituents cannot be reconstructed from today's membership. Intraday execution occurs in fragmented, scheduled markets with auctions, halts, tick sizes and venue routing. Short positions require locate and borrow, while an ETF adds fund-specific spreads, premiums/discounts and creation/redemption mechanics.

Equity capacity is highly cross-sectional. A strategy whose predictive power is concentrated in microcaps can show strong equal-weighted returns while supporting little capital. Value-weighting is not a complete solution because it changes the question; it can conceal whether a model has incremental information away from mega-cap beta. Studies should report performance by liquidity and market-cap segment, participation-rate sensitivity and the contribution of the smallest securities. Trading-cost evidence should be read conditionally: high turnover can destroy anomaly profits \citep{NovyMarx2016TransactionCosts}, yet actual institutional records show that some systematic styles remain scalable at meaningful size \citep{Frazzini2018TradingCosts}. The cost model must match the strategy and investor.

ETFs can reduce the security-selection and execution burden for macro or allocation agents. They offer liquid, diversified exposures and can make a generated sector or asset-class view implementable without hundreds of individual orders. But they also make benchmark discipline essential. Rotating among equity ETFs may largely time market beta, duration, size or geography; risk-adjusted value should be measured against a feasible static or volatility-managed allocation. Leveraged and inverse ETFs have path-dependent daily objectives, so a model that treats their ticker return as a scaled index return across longer horizons is misspecified.

\subsection{Centralized crypto spot}

Centralized crypto spot markets operate continuously across venues and quote currencies. They offer rapid data generation, public blockchain covariates and a large natural experiment in market fragmentation. Return studies find that momentum and investor attention predict cryptocurrency returns \citep{LiuTsyvinski2021RisksReturns}, and a factor structure involving market, size and momentum explains a substantial part of the cross section \citep{LiuTsyvinskiWu2022CommonRiskFactors}. Persistent cross-exchange price differences documented by \citet{MakarovSchoar2020CryptoArbitrage} show that limits to arbitrage can be economically important. These findings support the existence of structured signals and frictions; they do not imply riskless machine-extractable profit.

Crypto data are abundant but not automatically clean. Token histories are short and universes change through listings, delistings, forks, migrations and failures. A dataset assembled from surviving API symbols can omit precisely the assets that went to zero. Reported volume may be strategic rather than organic. \citet{Cong2023CryptoWashTrading} find extensive wash trading across less regulated exchanges, and \citet{AlooshLi2024BitcoinWash} provide direct evidence of wash trading in Bitcoin markets. A model trained on volume, order flow or social attention can learn venue manipulation as if it were investor demand. This may remain locally predictive, but it changes both interpretation and the risk that the behavior disappears after enforcement or venue redesign.

Centralized custody and counterparty exposure are part of net performance. An apparent arbitrage can require prefunded balances on several venues, exposing capital to default and withdrawal risk. Settlement delays and transfer limits prevent instantaneous convergence. Stablecoin quote assets introduce issuer and peg exposure. AI does not diversify these risks simply by monitoring more feeds; a portfolio layer must constrain venue and settlement concentration, and performance should charge capital tied up to make execution possible.

\subsection{Perpetual futures are not spot with leverage}

Perpetual futures have no fixed expiry and use funding or analogous transfers to keep the contract near an index. For position $q_t$, a simplified holding-period P\&L is

\begin{equation}
\mathrm{PnL}_{t,t+1}=q_t(P_{t+1}-P_t)-q_tP_t f_t-C_t-\mathrm{LiqLoss}_{t,t+1},
\end{equation}

where $f_t$ is the funding rate paid by a long when positive, $C_t$ covers trading costs, and liquidation loss is discontinuous rather than a small linear fee. Venue rules use index and mark prices, maintenance margin, insurance funds and auto-deleveraging mechanisms. A backtest based only on last-traded prices can therefore report a path that the margined account could not survive.

Crypto carry can be large precisely because it is not a free yield. \citet{Schmeling2026CryptoCarry} document substantial variation in crypto futures carry and connect high carry to scarce arbitrage capital, leverage constraints and crash risk. Earlier work on Bitcoin derivatives similarly highlights venue-specific pricing and microstructure \citep{Alexander2020BitMEX}. A strategy that earns funding by taking the other side of leveraged demand is exposed to basis moves, margin, exchange and tail risk; a directional agent that ignores funding can be right about spot direction and wrong about contract P\&L. Evaluation needs contract-level funding timestamps, mark-price liquidation simulation, collateral currency, leverage, margin mode, fee tier and downtime.

Perpetual datasets also create subtle leakage. A current contract specification may be applied to an earlier period with different funding intervals or limits. Continuous series can cross listing changes. The set of tradable tokens can be reconstructed using future liquidity. Liquidation labels derived from later-reported data can leak the event. A point-in-time contract master is as necessary as point-in-time fundamentals in equities.

\subsection{On-chain markets: transparent state, endogenous extraction}

Blockchain settlement makes balances, contract calls and transaction ordering visible, enabling features that centralized markets cannot expose. An agent can observe liquidity-pool state, flows between addresses, protocol positions and composable prices. The same transparency creates a competitive execution game. Pending or public transactions can be reordered or surrounded; users bid gas or priority fees; failed transactions consume resources; and contract state can change between simulation and inclusion. MEV is therefore not an operational footnote but a claimant on gross surplus.

AMMs replace or complement limit-order books with programmed liquidity. Liquidity providers earn fees while bearing inventory and adverse-selection losses. Loss-versus-rebalancing (LVR) formalizes the loss relative to an appropriate rebalancing benchmark \citep{Milionis2022LVR}. Empirical work finds that arbitrage-related losses can exceed fee income in many large Uniswap pools \citep{FritschCanidio2024LVR}. \citet{CapponiJia2025DEXLiquidity} show how arbitrage competition can transfer much of the surplus into gas and infrastructure expenditure, while \citet{CapponiJiaYu2026DEXPriceDiscovery} link fees and ordering behavior to price discovery. A model that predicts the next pool price can therefore be statistically excellent while failing to capture any surplus after priority competition.

For a swap of intended size $x$, executable output is a function of reserves at inclusion, fee tier, price impact and ordering, not simply the contemporaneous mid-price. For liquidity provision, the benchmark is not token buy-and-hold alone; fee income must be compared with rebalancing, hedging, gas and adverse selection. Smart-contract risk, oracle failure, bridge exposure and governance changes are additional state transitions. Backtests need block-level ordering assumptions, gas and failed-call accounting, realistic slippage bounds and a policy for reorg or finality. ``On-chain'' means observable after settlement, not necessarily known before a decision.

\subsection{A market-specific evaluation matrix}

Table~\ref{tab:market_matrix} prevents a common category error: applying one return equation and one cost scalar to all markets. Shared requirements include point-in-time universes, executable benchmarks, risk adjustment and capacity. The marginal cash flows and failure events differ.

\begin{table}[t]
\centering
\caption{Market strata require different P\&L and validity fields. Items in the final column are minimum additions to ordinary return reporting.}
\label{tab:market_matrix}
\scriptsize
\begin{tabularx}{\textwidth}{@{}p{2.05cm}Y Y Y@{}}
\toprule
Stratum & Distinct information/data risk & Distinct implementation loss & Minimum economic validation \\
\midrule
Listed equities/ETFs & Publication lags, revisions, constituent and delisting bias, corporate actions & Spread, impact, commission, borrow, auction/halts; ETF premium and path dependence & Liquidity/capacity slices, borrow and turnover, factor alpha, corporate-action-correct universe \\
Centralized crypto spot & Changing token/venue universe, wash volume, forks, quote-asset and API revisions & Fragmented fills, custody, prefunding, withdrawal and stablecoin risk & Venue-level executable prices, survival-free universe, custody/capital charge, manipulation sensitivity \\
Perpetual futures & Historical contract specifications, index/mark construction, changing funding rules & Funding, leverage, margin, liquidation, insurance/auto-deleveraging & Contract-level funding and mark-price path, collateral and margin rules, tail liquidation tests \\
On-chain/DEX & State at signing versus inclusion, address labels, reorg/finality, adversarial activity & Gas/priority fees, slippage, failed calls, MEV, LVR, smart-contract loss & Ordered transaction simulation, all failed/successful gas, rebalancing benchmark, contract/allowance controls \\
\bottomrule
\end{tabularx}
\end{table}

The matrix also explains why crypto warrants explicit treatment within a common validation framework rather than peripheral mention. It is not merely a volatile additional asset class on which to rerun an equity model. Its continuous trading and public state make it a valuable laboratory for adaptation and execution, while leverage, venue quality and programmable ordering expose failure modes more sharply. Evidence can transfer at the level of method---point-in-time data, cost-aware objectives, uncertainty and controls---but profitability numbers should remain stratified. The AI-specific crypto evidence examined here is thinner than the equity evidence; common criteria must not be mistaken for equal evidentiary depth.

%% file: sections/08_profitability_evidence.tex
\section{Does the Evidence Establish Sustainable Excess Returns?}
\label{sec:evidence}

\subsection{What the affirmative record establishes}

Within the reviewed record, credible affirmative evidence exists in several designs. Large-panel equity studies show that nonlinear ML can improve historical out-of-sample return prediction and portfolio outcomes \citep{Gu2020EmpiricalAssetPricing,Chen2024DeepAssetPricing}. Direct portfolio methods report economically meaningful utility or risk-adjusted gains relative to linear and forecast-first baselines \citep{Simon2026DeepPolicies,Cong2026AlphaPortfolio,Jensen2026ImplementableFrontier}. Text models extract return-relevant information from news under post-training-cutoff designs \citep{LopezLira2026ChatGPT}, and point-in-time language models produce economically meaningful signals without knowingly using later corpora \citep{Kelly2026PITLanguageModels}. Within those settings, the results contradict two categorical positions: that flexible models add no information, or that every favourable AI backtest is necessarily leakage.

One recent design also reports improvement on a selected upstream economic benchmark. \citet{KoijenLevy2026AgenticAssetPricing} construct a real-time out-of-sample task around earnings announcements and report that optimized agentic systems raise explained contemporaneous return variation from a standard benchmark near 8\% to close to 20\%. This is evidence, within that protocol, that agents can structure text into interpretable mechanisms. It is not a trading return: contemporaneous explanation around an announcement includes price adjustment that may occur before an executable position can be established. Its scientific value is strongest as a time-valid asset-pricing and information-processing benchmark.

The reviewed crypto studies likewise document structured opportunities rather than pure noise. Momentum, attention and cross-sectional factors appear in spot returns \citep{LiuTsyvinski2021RisksReturns,LiuTsyvinskiWu2022CommonRiskFactors}; persistent international price segmentation reveals limits to arbitrage \citep{MakarovSchoar2020CryptoArbitrage}; and futures carry is economically large in some states \citep{Schmeling2026CryptoCarry}. AI may improve the speed or conditionality with which such states are recognized. But these premia and mispricings can compensate exposure to crashes, constrained capital, custody and liquidation. Their existence is an input to an AI strategy, not evidence that the model captures the return net of competition.

\subsection{Why the examined record does not establish general profitability}

The evidence thins for four cumulative reasons. First, time validity is often incomplete. Historical LLM queries are especially unreliable when the base model has seen the outcome \citep{LopezLira2026Memorization}. More traditional ML is not immune: \citet{Zhang2025ManMachine} trace a reported monthly alpha to look-ahead alignment and find that removing it erases the alpha; the journal subsequently issued an expression of concern regarding the original article \citep{RFS2026ExpressionConcern}. These are not minor reproducibility details. They show that a plausible model can turn a data timestamp error into an economically large story.

Second, repeated search and publication selection inflate the observed maximum. Reality-check and backtest-overfitting methods formalize the problem \citep{White2000RealityCheck,Bailey2017BacktestOverfitting}. In an LLM or agent experiment the search space includes prompts, role descriptions, memory lengths, retrieval sources, model versions, random seeds, assets and windows. Reporting only the best trajectory understates uncertainty even if the final period is technically held out. A complete record should identify the frozen specification and distinguish exploratory validation from confirmatory testing.

Third, implementation and equilibrium erode signals. Costs are endogenous to turnover, urgency, liquidity and capital; cost-agnostic ML can select precisely the least scalable opportunities \citep{Jensen2026ImplementableFrontier}. After publication, anomaly returns decline in ways consistent with both initial overfit and investor learning \citep{McLeanPontiff2016Publication}. An AI strategy may accelerate its own decay: faster extraction makes the early price response more efficient, while common models and data reduce uniqueness. The falling strength of the headline signal as ChatGPT use diffuses in \citet{LopezLira2026ChatGPT} is consistent with this reflexive mechanism, although it does not identify all causes.

Fourth, emerging prospective and field evidence is more restrained than selected backtests. \citet{Carlin2026HouseholdAI} collect daily LLM stock recommendations prospectively. They find concentrated tilts toward large, growth and momentum stocks and media attention, while buy-and-hold and active AI portfolios do not earn statistically significant abnormal returns under their characteristic adjustment. The study is preliminary and its prompts, retail framing and period do not cover all professional systems; its importance is that it observes choices before outcomes and separates raw returns from style-adjusted skill. Using regulatory disclosures, labor-market information and strategy descriptions, \citet{Chen2026AIAssetManagement} find that AI hedge funds outperformed non-AI peers early, but that this advantage declined over time even among early adopters. They also find lower, not higher, return comovement among AI funds, qualifying the simple claim that AI necessarily homogenizes all strategies. Together these studies support potential and diffusion limits, not a zero-effect verdict.

\begin{table}[t]
\centering
\caption{Representative evidence and the maximum inference it supports. Numerical results are not compared across rows because universes, horizons, objectives and costs differ.}
\label{tab:profit_evidence}
\scriptsize
\begin{tabularx}{\textwidth}{@{}p{3.0cm}p{2.4cm}Y Y@{}}
\toprule
Evidence & Evaluation mode & Affirmative inference & Binding qualification \\
\midrule
Gu et al.; Deep SDF & Historical equity OOS & Nonlinear characteristic structure improves selected forecasts/portfolios & Historical, universe- and implementation-specific \\
Deep policies; AlphaPortfolio & Historical decision OOS & Economic objectives and sequence modeling can improve selected portfolios & Preference, search, constraint and cost-model dependence \\
Implementable frontier & Historical cost-aware OOS & Joint cost-aware learning dominates cost-agnostic construction in the study & Simulated costs and capacity; not live capital \\
Post-cutoff news LLM & Post-training-cutoff event test & LLM semantics can identify return-relevant news and later drift & Latency, adoption decay and portfolio implementation \\
Point-in-time LMs & Chronological model checkpoints & Useful text signals need not rely on future-trained models & Historical portfolio construction remains model-specific \\
Prospective household portfolios & Daily precommitted recommendations & Direct evidence of LLM style, concentration and behavior & No significant adjusted abnormal return; preliminary period \\
AI hedge-fund field data & Observed fund classifications and returns & Early AI adopters show potential; strategies are not simply homogeneous & Relative outperformance declines; AI classification is observational \\
Live/real-time agent benchmarks & Prospective paper or contemporaneous tasks & Reduce hindsight and test changing models & Generally short, small-scale, not audited live-capital alpha \\
\bottomrule
\end{tabularx}
\end{table}

\subsection{The bounded verdict}

Within the evidence examined here, the right conclusion is a gradient across profile dimensions. For representation, extraction and selected historical forecasts, evidence of progress is strong. For historical economic value under a disclosed portfolio and cost model, evidence is meaningful but heterogeneous. For prospective paper decisions, the reviewed record is growing and mixed. For independently audited live-capital performance that persists across regimes and capital scales, that public record is sparse. It therefore does not support a claim that a foundation-model, RL or agent family generally generates sustainable net alpha.

``Insufficient'' is not ``negative.'' A proprietary strategy may possess unique point-in-time data, careful execution and a record that cannot be published without destroying its edge. Nor would a universal public algorithm be economically stable if inexpensive replication immediately crowded it. The bounded claim is epistemic: the academic and public product evidence examined here cannot carry a stronger conclusion than its designs permit. For researchers, the implication is to strengthen each material evidence-profile dimension and use prospective or live designs when the claim requires them. For users, it is to ask which part of a return comes from market exposure, known styles, leverage, omitted cost and selected timing before attributing the residual to AI.

%% file: sections/09_research_agenda.tex
\section{A Research Agenda for More Credible and More Durable Net Returns}
\label{sec:agenda}

The reviewed evidence does not justify a recipe for alpha, but it identifies where research effort may improve durable economic value. The agenda is joint: better data cannot rescue a cost-blind objective, and a realistic simulator cannot rescue a contaminated model. Six linked directions follow from the alpha-translation chain.

\subsection{Make information and models point-in-time by construction}

Every observation should carry both an event time and an availability time. Fundamental databases need original vintages and restatements; universes need listings, delistings and membership at each date; news needs first-public timestamps and corrections; crypto needs historical symbol, contract and venue specifications; on-chain decisions need the state observable when a transaction was formed, not only after inclusion. Data lineage should connect each feature to its source and transformation. A compact ``information bill of materials'' can list providers, fields, lags, revisions, timezone, missing-value policy and universe rules.

The model itself belongs in that bill. Historical LLM evaluation should use point-in-time base checkpoints, not merely retrieval restricted to old documents. The checkpoint sequence of \citet{Kelly2026PITLanguageModels} demonstrates that such infrastructure is feasible at useful scale. When chronological pretraining is unavailable, researchers should prefer truly post-cutoff prospective tests and explicitly limit the inference, supplemented by memorization probes \citep{LopezLira2026Memorization}. Prompts, system instructions, tool schemas, embedding models and external search indices must be versioned because any can encode later information.

This standard will initially reduce headline performance and sample length. That is a feature: it converts hidden bias into visible uncertainty. Shared, versioned point-in-time testbeds across equities and each crypto stratum would create more cumulative knowledge than another untraceable leaderboard.

\subsection{Train for economic decisions, including abstention}

Targets should reflect the action and horizon. Next-period squared error is useful when the output genuinely feeds a mean forecast; it is poorly aligned with ranking, event surprise, tail risk or execution. Decision-focused objectives can include risk, turnover and costs directly, as the implementable-frontier and deep-policy evidence illustrates \citep{Jensen2026ImplementableFrontier,Simon2026DeepPolicies}. Distributional forecasts are preferable when position size depends on uncertainty, and calibrated abstention should be an explicit action. A model that declines low-quality trades may create more value than one forced to issue a daily buy/sell label.

Economic objectives must be evaluated for specification gaming. Researchers should show which holdings and market states produce gains; perturb risk aversion, cost and financing; and compare against simple policies using the same information. Multi-objective reporting is safer than optimizing one reported Sharpe ratio. At minimum, return, risk, turnover, concentration, capacity and tail loss should remain visible rather than collapsed into a score whose weights were selected after the test.

\subsection{Co-design signal, portfolio, and execution}

The usual modular pipeline freezes a forecast before learning whether it is tradable. A better design passes cost and capacity information upstream. Slow signals may be aggregated and netted; fast signals may justify trading only in liquid names or at sufficiently high confidence; execution uncertainty may reduce target positions. Portfolio combinations can cancel trades rather than simply average forecasts \citep{DeMiguel2020TransactionCostManaged}. Execution policies should optimize a fixed parent decision and be measured by implementation shortfall, keeping signal quality separate from fill quality.

Capacity curves should be first-class results. Report net performance over capital, participation rate and latency, with spread, nonlinear impact, borrow/funding and venue-specific losses varied independently. For crypto, use separate engines for spot, perpetual and on-chain P\&L. An agent that chooses a perpetual must receive the funding and liquidation state; one that calls a DEX needs transaction-order and failed-gas simulation. Cross-venue arbitrage requires a capital and counterparty model, not instantaneous transfer at displayed prices.

Market impact creates a further research opportunity: train policies on endogenous response rather than a fixed tape. Historical limit-order-book replay can support small-action evaluation, while calibrated agent-based or structural simulators can explore counterfactual larger actions. Because simulator misspecification is inevitable, policies should be stress-tested across several plausible response models and graduate through paper, shadow and limited live stages. Robustness to a family of simulators is more informative than mastery of one.

\subsection{Treat adaptation as a controlled experiment}

Financial relationships decay through macro change, regulation, venue redesign, competition and the model's own adoption. Adaptation is therefore necessary, but unconstrained online updating makes evaluation and governance ambiguous. A deployable system needs predeclared triggers, candidate generation, shadow comparison, approval, rollback and an immutable record of the active version. Performance monitoring should attribute changes to beta, style, volatility, liquidity, costs, model output and execution rather than watching cumulative return alone.

Research can distinguish several adaptation mechanisms: rolling or expanding retraining; state-conditioned mixtures; change-point detection; meta-learning; memory retrieval; and human revision. They should be compared under the same information and compute budget. A memory-equipped LLM may appear adaptive because it reads recent outcomes, while a simple volatility-conditioned rule may explain the gain. Tests should include structural breaks not used to tune the detector, source outages, asset entry and exit, fee-rule changes and adversarial news.

Crowding should also enter evaluation. Post-publication decay \citep{McLeanPontiff2016Publication} and declining AI-fund relative performance \citep{Chen2026AIAssetManagement} imply that a static historical edge is not a stable resource. A useful model estimates how return and capacity change as similar capital grows. Research consortia can test this indirectly through signal overlap, trade correlation and response to public release, while recognizing that exact proprietary positions will remain unavailable.

\subsection{Replace retrospective showcases with prospective evidence}

The next generation of benchmarks should require a precommitted decision artifact before the outcome. A public append-only ledger can record model and data versions, eligible universe, prompt/policy hash, target holdings, order timestamp and subsequent correction policy. Paper portfolios should use executable bid/ask or documented fill models and preserve rejected or missing decisions. Model updates should start a new cohort rather than rewriting the prior record. Prospective household recommendations \citep{Carlin2026HouseholdAI}, real-time earnings benchmarks \citep{KoijenLevy2026AgenticAssetPricing} and live-market agent arenas \citep{Qian2026AgentMarketArena} show complementary ways to reduce hindsight.

Prospective evaluation does not require immediate unrestricted capital. A safe sequence is offline frozen test, historical event replay, forward shadow decisions, broker paper orders, small-capital controlled deployment and only then scaled authority. Each stage answers a different question. A study can be valuable if it stops at paper trading, provided it does not describe simulated fills as live trading. Independent replication should reproduce the decision log and metric calculation, not merely rerun open code on a newly downloaded and potentially revised dataset.

Statistical protocols must count search. Pre-registering one final specification is ideal but not always compatible with systems research. An alternative is a transparent model-development ledger, a sequestered confirmatory period and family-wise or false-discovery control across material variants \citep{Harvey2016FactorZoo,White2000RealityCheck}. Uncertainty should reflect serial dependence and stochastic model runs. A table of all attempted configurations is often more scientifically informative than an additional best-case chart.

\subsection{Engineer governance in proportion to authority}

Governance is part of attainable performance because operational loss can overwhelm a statistical edge. The NIST AI Risk Management Framework provides general functions for governing, mapping, measuring and managing AI risk \citep{NIST2023AIRMF}; IOSCO documents capital-market AI use cases and risks from a securities-regulatory perspective \citep{IOSCO2025AICapitalMarkets}. Existing market-access controls also supply a concrete principle: systems with order authority require pre-trade financial and regulatory limits, controlled access, monitoring and review \citep{SEC2011MarketAccess}. These sources do not certify a particular trading agent, nor do they substitute for jurisdiction-specific legal analysis. They anchor a control architecture.

A practical separation has four planes. The \emph{information plane} retrieves untrusted external content and must defend against prompt injection, stale data and malicious files. The \emph{decision plane} estimates signals and target positions but cannot alter limits. The \emph{execution plane} validates identity, price, notional, leverage, venue, instrument and freshness through deterministic controls. The \emph{assurance plane} independently records inputs, approvals, orders, fills, reconciliations, incidents and model versions. Credentials are scoped to the minimum action, stored outside prompts, rotated and revocable. Natural-language confirmation is not a substitute for structured order validation.

For LLM agents, reproducibility and semantic risk are linked. Generated rationales can cite nonexistent evidence, confuse units or change a ticker through an ambiguous name. Retrieval should return source identifiers and timestamps; numerical quantities should be parsed into typed schemas and checked by deterministic code; conflicting sources should remain visible; and the system should fail closed when instrument identity or unit conversion is uncertain. Human oversight must be defined by an actual decision right and response window, not the statement that a human is ``in the loop.''

The Federal Reserve's revised model-risk guidance illustrates the importance of reading governance scope precisely: its 2026 supervisory letter revises expectations for models used by banking organizations and states that generative and agentic AI fall outside the guidance's defined model scope, even though broader risk-management obligations may still apply \citep{FederalReserve2026ModelRisk}. It would therefore be wrong to cite that letter as direct approval or a complete rulebook for trading agents. The general lesson is to map each system to the rules governing its institution, activity, data and market access rather than invoke ``AI compliance'' abstractly.

\subsection{A falsifiable program, not an optimism premium}

These directions should be evaluated as hypotheses. Point-in-time pretraining may reduce apparent performance without finding a replacement edge. Cost-aware objectives may overfit a cost curve. Joint execution learning may add complexity without improving fills. Governance may reduce speed while preventing rare losses. The appropriate outcome is a set of measured trade-offs. A credible research program should publish negative results, document where the alpha-translation chain breaks and distinguish failure of a signal from failure of its implementation.

The most promising competitive advantage may be less visible than a new model name: cleaner proprietary timestamps, a unique but lawful information source, faster error correction, lower turnover, better risk aggregation, patient execution and disciplined decision rights. Foundation models and agents can amplify these capabilities, but they can also commoditize public analysis. The synthesis developed here suggests that sustainable net alpha, where it exists, is more plausibly a property of the complete socio-technical investment process than of one architecture alone.

%% file: sections/10_conclusion.tex
\section{Conclusion}
\label{sec:conclusion}

The reviewed evidence documents material technical progress at several points in the investment chain. Nonlinear models extract structure from broad equity panels; deep and reinforcement-learning policies place economic objectives inside portfolio construction; time-series models transfer representations; language models process news and disclosure at scale; and agents combine retrieval, memory, analysis, tools and execution interfaces. Crypto markets extend the opportunity set through continuous data and observable blockchain state while exposing funding, liquidation, manipulation, custody, gas, MEV and smart-contract risks that cannot be represented by an equity-style cost scalar.

The public profitability record examined here is narrower than this technical record. Strong historical out-of-sample findings coexist with anomaly decay, multiple-testing risk, timestamp corrections that erase returns, LLM memorization, cost- and capacity-sensitive frontiers, prospective portfolios without significant abnormal returns, and limited independently audited live-capital evidence. The resulting conclusion is not that AI cannot earn excess returns. It is that the evidence examined through 31 August 2026 does not establish a general AI architecture that produces persistent, cross-regime, capacity-aware net alpha.

Progress should therefore be judged by the weakest link between information and realized net performance. Point-in-time data and model versions, decision-aligned objectives, joint portfolio--execution design, explicit market-specific P\&L, controlled adaptation, prospective records and authority-matched governance can make future claims more credible and may improve economic outcomes. None guarantees profit. Their value is that they convert an attractive backtest into a sequence of falsifiable claims---and make failure visible before it becomes uncontrolled capital loss.

%% file: sections/11_ai_disclosure.tex
\section*{Declaration of Generative AI Use}

Generative AI tools assisted with literature discovery, drafting, language editing, and LaTeX preparation; the authors verified the sources, interpretations, and final text and take full responsibility for the manuscript.